\pdfoutput=1

\documentclass[11pt]{article}

\usepackage[preprint]{acl}

\usepackage{times}
\usepackage{latexsym}

\usepackage[T1]{fontenc}

\usepackage[utf8]{inputenc}

\usepackage{microtype}

\usepackage{inconsolata}

\usepackage{graphicx}
\usepackage{xspace}

\usepackage{adjustbox}
\usepackage{makecell}
\usepackage{multirow} 
\usepackage{array}
\usepackage{soul}
\usepackage{caption}
\usepackage{subcaption}
\usepackage{enumitem}
\newcommand{\method}{\textsc{RwG}\xspace }

\title{Reasoning with Graphs: Structuring Implicit Knowledge to Enhance LLMs Reasoning}

\author{
    Haoyu Han\textsuperscript{1}, 
    Yaochen Xie\textsuperscript{2}, 
    Hui Liu\textsuperscript{2,1}, 
    Xianfeng Tang\textsuperscript{2}, 
    Sreyashi Nag\textsuperscript{2}, \\
    \textbf{William Headden\textsuperscript{2}, 
    Yang Li\textsuperscript{2}, 
    Chen Luo\textsuperscript{2}, 
    Shuiwang Ji\textsuperscript{3}, 
    Qi He\textsuperscript{2}, 
    Jiliang Tang\textsuperscript{1}} \\
    \textsuperscript{1}Michigan State University,
    \textsuperscript{2}Amazon,
    \textsuperscript{3}Texas A\&M University \\
    \texttt{\{hanhaoy1, liuhui7, tangjili\}@msu.edu, {sji@tamu.edu}} \\
    \texttt{\{yaochx, liunhu, xianft, sreyanag, headdenw, limyng, cheluo, qih\}@amazon.com} 
}

\begin{document}
\maketitle

\begin{abstract}
Large language models (LLMs) have demonstrated remarkable success across a wide range of tasks; however, they still encounter challenges in reasoning tasks that require understanding and inferring relationships between distinct pieces of information within text sequences. This challenge is particularly pronounced in tasks involving multi-step processes, such as logical reasoning and multi-hop question answering, where understanding implicit relationships between entities and leveraging multi-hop connections in the given context are crucial. Graphs, as fundamental data structures, explicitly represent pairwise relationships between entities, thereby offering the potential to enhance LLMs' reasoning capabilities. External graphs have proven effective in supporting LLMs across multiple tasks. However, in many reasoning tasks, no pre-existing graph structure is provided. Can we structure implicit knowledge derived from context into graphs to assist LLMs in reasoning?
In this paper, we propose Reasoning with Graphs (RwG) by first constructing explicit graphs from the context and then leveraging these graphs to enhance LLM reasoning performance on reasoning tasks. 
Extensive experiments demonstrate the effectiveness of the proposed method in improving both logical reasoning and multi-hop question answering tasks.
\end{abstract}

\section{Introduction}
\label{sec:intro}
\begin{figure}[!htb]
    \centering
   \includegraphics[width=\linewidth]{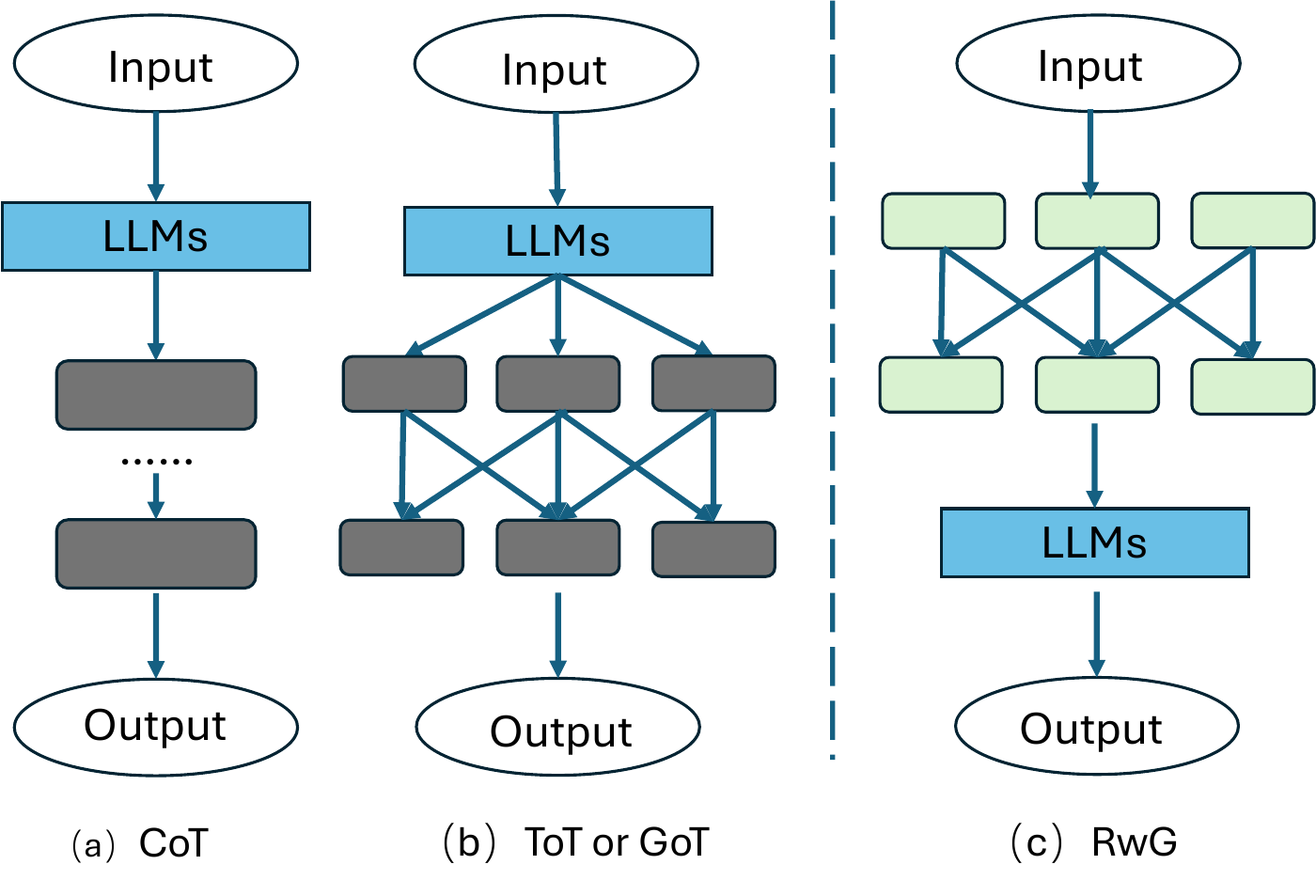}
    \caption{Comparison of Reasoning with Graph (RwG) to other prompting methods.  Gray nodes represent thoughts generated by LLMs, while green nodes represent entities extracted from the input.}
    \label{fig:intro}
    \vspace{-0.2in}
\end{figure}

In recent years, Large Language Models (LLMs) have showcased remarkable capabilities across a variety of tasks, such as question answering~\citep{zhuang2024toolqa, lan2022complex}, summarization~\citep{pu2023summarization}, and language understanding~\cite{zhao2023survey}. Despite these successes, LLMs still encounter significant challenges in certain areas~\cite{zhao2023survey, minaee2024large}. A key limitation lies in their struggle with reasoning tasks~\citep{yang2024large, huang2023large}, particularly with logical reasoning~\citep{nezhurina2024alice}, which requires models to infer missing relationships between distinct pieces of information, and multi-hop reasoning~\citep{yang2024large}, where they must trace a path or follow some structures through the context to arrive at the correct answer.

To enhance the reasoning capabilities of LLMs, several prompting methods have been proposed. Chain-of-Thought (CoT) prompting~\cite{wei2022chain, kojima2022large} aids LLMs in reasoning by generating intermediate steps that lead to the final answer. CoT has shown significant improvement in certain reasoning tasks without requiring model tuning. Building on CoT, the Self-Consistency method~\citep{wang2022self} further enhances reasoning by generating multiple CoT pathways and selecting the most consistent one. Additionally, Tree of Thought (ToT)~\citep{yao2024tree} and Graph of Thoughts (GoT)~\cite{besta2024graph} extend this approach by structuring the LLMs' thought process using trees and graphs, respectively. These methods prompt LLMs to generate initial thoughts and organize them into various structures. Despite the successes of these approaches, they still face challenges in handling complex reasoning tasks, such as logical reasoning~\citep{nezhurina2024alice} and multi-hop question answering~\citep{yang2024large}. 

For these complex reasoning tasks, LLMs need to figure out the relationships between entities in the context and infer missing components. Take the AIW+ problem ~\citep{nezhurina2024alice} as one example, LLMs are asked to solve problems such as \textit{Alice has 3 sisters. Her mother has 1 sister who does not have children - she has 7 nephews and nieces and also 2 brothers. Alice’s father has a brother who has 5 nephews and nieces in total, and who has also 1 son. How many cousins does Alice’s sister have?} In this problem, LLMs need to infer the relationships between each character, such as the relation between Alice and her mother's sister. Additionally, LLMs must infer the missing roles in the question, such as identifying the nephews and nieces that Alice's father's brother has. These types of questions pose significant challenges for LLMs, with many popular models achieving nearly zero accuracy on these tasks~\citep{nezhurina2024alice}. Typically, LLMs treat the information as a sequence. However, if a human were asked to solve this problem, a natural approach would be to draw a graph representing the relationships between each character and infer the missing links based on that graph. This is because graphs provide a fundamental data structure for representing relationships between entities, and the structural information is crucial for reasoning tasks.

Several works have shown the effectiveness of leveraging external graphs to help LLMs in reasoning, such as improving retrieval quality using graph structures~\citep{he2024g, tian2024graph} or reasoning on an external graph~\citep{jin2024graph, luo2023reasoning}. However, these methods rely on pre-existing graph structures. In most common reasoning tasks, only textual sequences are available. Therefore, a natural question arises: ``Can LLMs enhance their reasoning abilities by structuring implicit knowledge into explicit graphs?"

In this work, we aim to explore reasoning with graphs by constructing explicit graph structures from the context. Unlike previous prompting approaches, which construct trees or graphs based on LLMs' thoughts, our Reasoning with Graphs (\method) method directly constructs explicit graphs from the context, where nodes are the entities in the context. The comparison is shown in Figure~\ref{fig:intro}.
Specifically, we first design a graph construction method with multiple rounds of verification to generate a graph from the given context for the reasoning problem. We then assess the LLMs' reasoning abilities with the constructed graph. Experimental results demonstrate that the proposed \method significantly improves the performance of various LLMs on both logical reasoning and multi-hop question answering tasks. \method showcases the potential of leveraging explicit graph structures derived from the context to enhance LLM reasoning capabilities, offering a promising new direction for incorporating structured knowledge into LLM-driven tasks.



\section{Related Works}
\label{sec:relate}
\subsection{Reasoning of Large Language Models}

Reasoning is a fundamental aspect of human intelligence, crucial for problem solving, decision making, and critical thinking.  Recent advancements in LLMs, such as GPT-4~\citep{achiam2023gpt} and LLaMA-3~\citep{touvron2023llama}, suggest that the ability for reasoning is already embedded within these large-scale models. Various prompting methods have been proposed to better utilize the reasoning capabilities of LLMs. Chain-of-Thought (CoT)~\citep{wei2022chain} is one of the most popular methods, prompting LLMs to generate reasoning paths. Building on this concept, Tree-of-Thought (ToT)~\citep{yao2024tree} and Graph-of-Thought (GoT)~\citep{besta2024graph} similarly model different reasoning paths using tree or graph structures. In addition to designing prompts, adopting additional strategies, such as incorporating verifiers, has contributed to enhancing the reasoning abilities of large language models. For instance, self-consistency~\citep{wang2022self} improves LLMs' reasoning by using majority voting among multiple generated paths. Studies by \cite{weng2022large} and \cite{stechly2024self} demonstrate that LLMs can benefit from self-verification or external verification methods.
Additionally, other techniques have been introduced to enhance LLMs' reasoning abilities, such as in-context learning~\citep{lampinen2022can}, fine-tuning~\citep{rajani2019explain}, and retrieval-augmented generation (RAG)\citep{huang2022towards, qiao2022reasoning, gao2023retrieval}. Recent studies~\citep{wang2024chain} reveal that CoT reasoning paths can be elicited from pre-trained LLMs simply by altering the decoding process without explicit prompting. This demonstrates that the effectiveness of CoT lies in guiding LLMs toward different decoding paths; for example, CoT can choose longer and more reliable paths instead of relying on greedy decoding. In this paper, we explore a different approach to prompting LLMs' reasoning abilities. Rather than leveraging multiple generated thoughts, we model the reasoning tasks as graphs, where the nodes represents entities in the question, and test the LLMs' ability to reason directly with these graph structures.

\subsection{Graphs for LLMs}

Graphs, which represent relationships between entities, are popular data structures widely used across various domains~\citep{ma2021deep}. Recently, numerous studies have explored the integration of graphs with LLMs. While LLMs have shown promise in assisting with certain graph-related tasks~\citep{jin2023large, chen2024exploring}, several evaluations~\citep{guo2023gpt4graph, liu2023evaluating} have revealed that LLMs often struggle to understand basic graph concepts, such as degree, shortest path, and motifs, particularly when dealing with large graphs. Additionally, several works have sought to enhance LLMs' reasoning abilities using graphs by retrieving relevant information. These methods typically involve extracting ego subgraphs based on related nodes and edges~\citep{zhang2022greaselm, tian2024graph} or paths within knowledge graphs~\citep{luo2023reasoning}. Furthermore, GraphReason~\citep{cao2023enhancing} constructs a graph based on LLMs' outputs and then verifies the output using the graph. However, these methods rely on external graphs or generate graphs based on LLMs' reasoning paths; they do not explore the effects of directly constructing a graph from the reasoning problems. Two related approaches are worth mentioning: GE-Reasoning~\citep{park2023graph}, which decomposes multi-hop questions into sub-questions to form a graph and prompts LLMs to answer based on the chronological order of the graph and Structure-Guided Prompting~\citep{cheng2024structure}, which builds a graph from text to solve graph-based tasks. In this paper, we construct graphs from the context of complex reasoning questions and use these graphs to assist LLMs in their reasoning processes.

\section{Reasoning with Graph}
\label{sec:method}

Many reasoning tasks involve inferring missing entities and relationships that are not explicitly presented in the question. Graphs provide an explicit structure to represent relationships between key entities and serve as a useful tool for inferring missing connections. However, reasoning problems typically do not come with explicit graph representations. Reasoning with Graph (\method) teaches large language models to tackle complex reasoning questions by structuring the implicit knowledge within the questions into explicit graph representations and leveraging these graph structures to solve the problems. This mirrors how humans often solve complex reasoning problems — by organizing information in a structured way, such as drawing diagrams to clarify connections between concepts. In \method, no additional or external graph information is used; the graph serves solely as a tool for reasoning. 

We roughly decompose the process of proposed \method into two key stages: (1) \textbf{Graph Construction}: The graph construction prompt guides LLMs to build an explicit graph based on the context of the reasoning question. We expect the graph to meet different requirements depending on the tasks, which are detailed in sections~\ref{sec:experiments}. (2) \textbf{Reasoning with graph}: Once the graph is constructed, the reasoning question is answered by leveraging the information encoded in the graph structure. Next, we will provide a detailed explanation of each stage.

\begin{figure*}[htb]
    \centering
   \includegraphics[width=\linewidth]{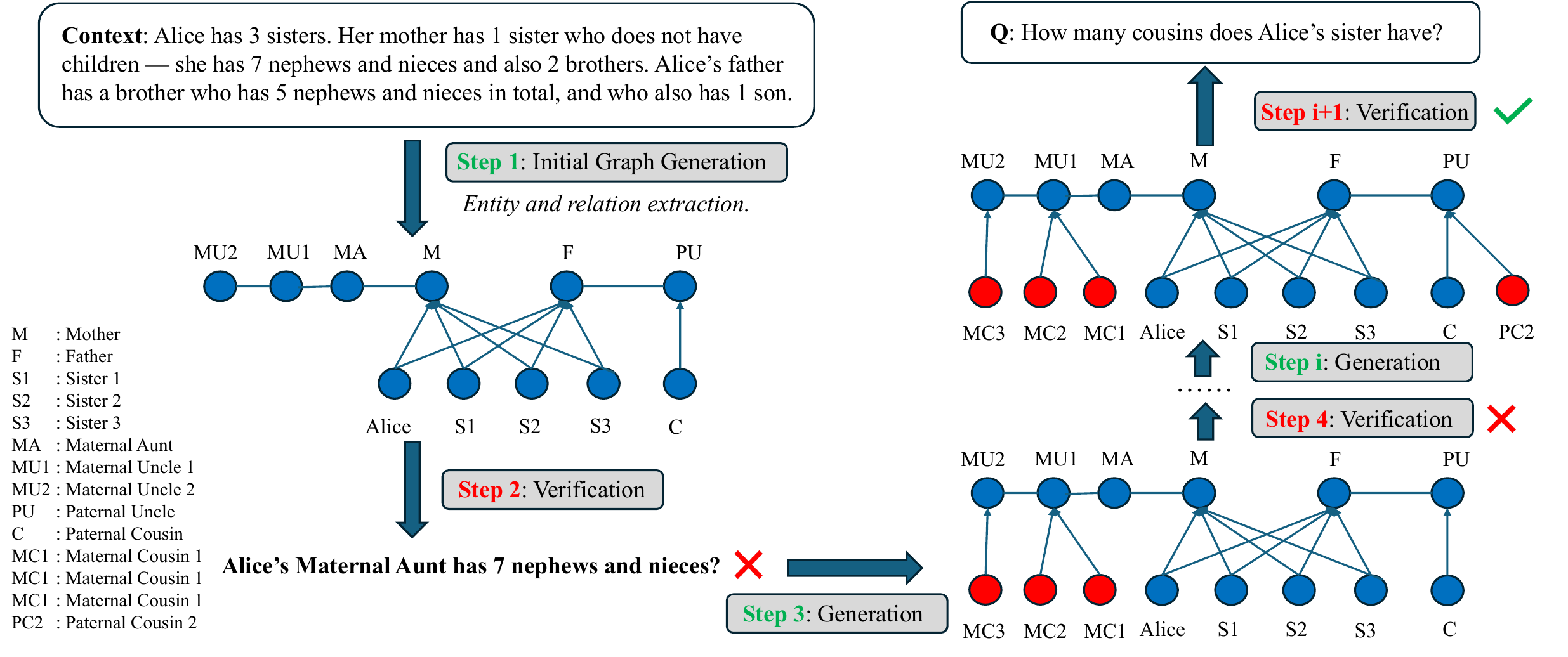}
    \caption{The procedure of \method for the AIW+ example. Blue nodes represent entities explicitly mentioned in the context and included in the initial graph, while red nodes denote inferred entities added during the graph generation and verification processes. The node names are based on their relationship to Alice.}
    \label{fig:graph}
    \vspace{-0.2in}
\end{figure*}

\subsection{Graph Construction}

The goal of this step is to construct a graph from unstructured reasoning problems, representing the relationships between the entities mentioned in the reasoning question.
However, there may be missing entities or relationships that are not explicitly stated in the context. These missing elements could be critical to solve the reasoning question. Therefore, we should refine the constructed graph by inferring additional relationships or entities, ensuring that it satisfies the requirements of the reasoning problem.

Take the context of AIW+ problem as an example: \textit{Alice has 3 sisters. Her mother has 1 sister who does not have children—she has 7 nephews and nieces and also 2 brothers. Alice’s father has a brother who has 5 nephews and nieces in total, and who also has 1 son}. In the constructed graph, all the characters mentioned in the context, such as Alice, and Alice's father's brother should be included. Additionally, it should include missing roles and relationships. For instance, Alice and her sisters account for only 4 of the 5 nephews and nieces of her father's brother, implying that there is one missing individual, which should be included in the graph.

There are several traditional methods for graph construction, such as entity and relation extraction~\citep{zhong2023comprehensive}. With the advancements in LLMs, recent works~\citep{edge2024local, zhang2024extract} have also leveraged LLMs to automatically detect entities and relationships for graph construction. These methods can be used to generate an initial graph for reasoning questions.

However, the constructed initial graphs may only capture the entities and relationships explicitly mentioned in the context, which may not fully meet the requirements for different tasks. For example, in the logical reasoning tasks, there might be some constraints in the context, such as ``Alice's father's brother has 5 nephews and nieces". In multi-hop question-answering tasks, crucial relationships between entities may be missing, which are essential for reasoning. To address this, we propose an iterative graph construction method that updates the graph repeatedly to meet the specific requirements for different tasks.
Specifically, this process mainly involves two steps: graph generation and graph verification. The graph generation step aims to construct a graph based on the context, previous graph and feedback from the verifier. The graph verification step verifies whether the generated graph meets the requirements. 

We begin by prompting the LLMs to generate an initial graph based on the given query. Next, we ask the LLMs to verify whether the graph satisfies the requirements. If the graph does not meet the requirements, the LLMs are prompted to add the missing entities or relations to update the graph. This process of graph verification and graph generation is repeated until the graph satisfies all the requirements or the maximum number of iterations is reached. After construction, the graph is represented as a list of triples, with each triple consisting of a (Head Entity, Relation, Tail Entity).

The process of constructing a graph for the AIW+ example is illustrated in Figure~\ref{fig:graph}. We first prompt the LLMs to extract entities and relationships from the context to generate an initial graph. The blue nodes represent entities explicitly stated in the context and are shown in the initial graph, while the red nodes are inferred during the multiple rounds of graph verification and generation. During the graph generation and verification steps, LLMs can better understand the context and infer missing relationships by utilizing the explicit graph structure. Once the graph is complete, it can then be used to answer the questions.

\subsection{Reasoning with graphs}
Reasoning with the graph involves answering reasoning questions using both the constructed graph and the given context. Many existing methods leverage external graphs, such as training a graph-based encoder~\cite{tian2024graph, zhang2022greaselm}, retrieving subgraphs~\cite{he2024g, zhang2022subgraph}, or reasoning along a path within the graph~\cite{luo2023reasoning, sun2023think}. The constructed graph can also be utilized in these ways. However, in this paper, we focus on having LLMs directly solve the reasoning question by leveraging the graph and context. We prompt LLMs to answer reasoning questions based on the constructed graph and context. Additional approaches to utilizing the constructed graph can be explored as future work.


\section{Experiment}
\label{sec:experiments}
Our framework is inherently task-agnostic, designed to accommodate a wide range of tasks with versatility. To evaluate whether the proposed \method approach can enhance LLMs' reasoning and grounded generation capabilities, we test it on two distinct reasoning tasks: logical reasoning and multi-hop question answering. In all experiments, we follow a zero-shot setting.
\begin{table*}[htb]
\caption{The results on the AIW and AIW+ datasets. Since the AIW+ dataset contains many possible relationships, there are no \method + Relation results.}
\label{tab:aiw}
\resizebox{\textwidth}{!}{%
\begin{tabular}{c|cccc|cccc}
\hline
Datasets & \multicolumn{4}{c|}{AIW} & \multicolumn{4}{c}{AIW+} \\ \hline
Methods & Claude & GPT-4o & Llama3.1-8B & Llama3.1-70B & Claude & GPT-4o & Llama3.1-8B & Llama3.1-70B \\
Vanilla & \textbf{0.026} & 0.066 & 0.053 & 0.013 & 0.0588 & 0.1176 & 0 & 0.2941 \\
CoT & 0.013 & 0.5733 & \textbf{0.066} & 0.053 & 0 & 0.2352 & \textbf{0.058} & 0.3529 \\
Self-Consistency & 0 & 0.053 & 0 & 0 & 0.0588 & 0.0588 & 0 & 0.1176 \\
Least-to-Most & 0 & 0.4533 & 0.0266 & 0.053 & 0 & 0.1764 & 0 & 0.1764 \\
\method & \textbf{0.026} & 0.6266 & 0 & 0.12 & \textbf{0.2941} & \textbf{0.5294} & 0 & \textbf{0.4545} \\
\method + Relation & \textbf{0.026} & \textbf{0.8666} & 0.0266 & \textbf{0.5733} & - & - & - & - \\ \hline
\end{tabular}
}
\end{table*}

\begin{table*}[htb]
\caption{The results on LogiQA and AR-LSAT datasets.}
\label{tab:logiqa}
\resizebox{\textwidth}{!}{%
\begin{tabular}{c|cccc|cccc}
\hline
Datasets & \multicolumn{4}{c|}{LogiQA} & \multicolumn{4}{c}{AR-LSAT} \\ \hline
Methods & Claude & GPT-4o & Llama3.1-8B & Llama3.1-70B & Claude & GPT-4o & Llama3.1-8B & Llama3.1-70B \\
Vanilla & 0.387 & 0.5698 & 0.1827 & 0.5698 & 0.2565 & 0.3608 & 0.1217 & 0.313 \\
CoT & 0.3978 & 0.5483 & 0.3548 & 0.5053 & 0.213 & 0.3565 & 0.1782 & 0.2434 \\
Self-Consistency & 0.3871 & 0.5806 & 0.172 & 0.5483 & 0.2608 & 0.3521 & 0.1217 & 0.2826 \\
Least-to-Most & 0.3225 & 0.5806 & 0.2795 & 0.5483 & 0.2652 & 0.3565 & 0.1695 & 0.2695 \\
\method & \textbf{0.4516} & \textbf{0.6344} & \textbf{0.3871} & \textbf{0.5913} & 0.2782 & 0.4043 & 0.1826 & 0.3173 \\
\method + Self-Consistency & 0.4408 & 0.6451 & 0.3548 & 0.5591 & \textbf{0.3086} & \textbf{0.4521} & \textbf{0.2086} & \textbf{0.3217} \\ \hline
\end{tabular}
}
\end{table*}

\subsection{Task1: Logical Reasoning}

Logical reasoning is a crucial aspect of human reading comprehension and question answering.  A typical logical reasoning problem consists of a paragraph of facts and a question that requires the testee to draw a valid conclusion based on those facts. To generate a correct answer, a machine must not only understand the facts but also recognize the relationships between the different components in the question. By constructing a graph for the logical reasoning question, we explicitly extract the key entities and their relationships, while also inferring any missing entities and relations — an essential step for effective logical reasoning. The general generation and verification process for logical reasoning task in \method is as follows: (1) \textbf{Generation}: Generate a graph based on the context by updating the previous graph, inferring missing entities and relations; (2) \textbf{Verification}: Verify whether the graph meets all requirements outlined in context.

\subsubsection{Datasets}

We selected four popular logical question answering datasets: AIW, AIW+\citep{nezhurina2024alice}, LogiQA~\citep{liu2020logiqa} and AR-LSAT~\citep{wang2022lsat}. Specifically, the AIW and AIW+ datasets mainly focus on answering questions related to Alice and her family. The LogiQA dataset includes various types of reasoning questions. The AR-LSAT dataset is a complex logical reasoning dataset that tests the ability to analyze a scenario governed by a set of constraints and determine which option satisfies or conflicts with those constraints. For these datasets, where the answers are numbers or options, we use accuracy as the evaluation metric. For more details on these datasets and pre-processing, please refer to Appendix~\ref{app:logical_dataset}.

\subsubsection{Baselines}

We evaluate our method on four widely used LLMs: GPT-4o~\citep{achiam2023gpt}, Claude 3-sonnet~\citep{anthropic2024claude}, LLaMA3.1 8B, and LLaMA3.1 70B~\citep{touvron2023llama}. Additionally, we compare our results with several representative baselines, such as the Chain-of-Thought (CoT)~\citep{wei2022chain}, Least-to-Most Prompting~\citep{zhou2022least}, and Self-Consistency~\cite{wang2022self}. For the AIW dataset, there are only 3 different relations between entities, i.e., brother-brother, brother-sister, sister-sister. We introduce a variant of \method called \method-Relation, where explicit relationships are provided during the graph generation process. For \method, we set the maximum number of graph generation and verification steps to 5. The prompts in the proposed \method for these dataset are shown in Appendix~\ref{app:prompt}.

\subsubsection{Results}
The results of different LLMs on the AIW and AIW+ datasets are presented in Table~\ref{tab:aiw}, while the results for the LogiQA and LSAT datasets are shown in Table~\ref{tab:logiqa}. From these results, we can make the following observations:
\begin{itemize}[leftmargin=0.2in]
\setlength\itemsep{-0.2em}
\item Reasoning with graphs (\method) generally enhances the logical reasoning capabilities of various LLMs on all datasets. 
\item  Stronger models, such as GPT-4 and LLaMA 70B, tend to benefit more from \method. However, if the original model struggles to solve the problem, as seen with LLaMA 3.1-8B on the AIW and AIW+ datasets, applying \method usually does not yield significant improvements. 
\item When explicit relationships are provided, as in \method-Relation for the AIW problem, the reasoning ability is further enhanced. 
\item The proposed \method can be incorporated with other methods, such as Self-Consistency, and combining these approaches may achieve even better results, such as on AR-LSAT dataset.
\end{itemize}

\subsubsection{Case studies}
\label{sec:case}
To understand why the proposed \method improves performance on logical reasoning questions, we conduct case studies. Detailed results and additional examples are provided in Appendix~\ref{app:case}. Here, we analyze the behavior of \method with GPT-4o using the example shown in Figure~\ref{fig:graph}. The step-by-step procedure of \method is presented in Table~\ref{tab:alice}. In the first phase, \method generates an initial graph with the explicit entities mentioned in the question. The graph is then updated if it fails verification. For example, \method adds more entities to the maternal and paternal parts during the first and second rounds of verification, respectively. In the third round, \method identifies an incorrect relation from the second round and corrects it, successfully passing verification. Finally, the LLMs can answer the question correctly based on the complete graph. For other baselines, the LLMs may fail due to incomplete information, as demonstrated in Appendix~\ref{app:case1}.

\begin{table}[!htb]
  \centering
    \caption{The graph updating procedure of \method applied to the AIW+ example using GPT-4o.}
    \label{tab:alice}
    \resizebox{\linewidth}{!}{
\begin{tabular}{l}
    \hline
    \textbf{Initial Graph} \\
    \hline
    Alice's Mother -- Alice \\
    Alice's Mother -- Alice's Sister 1 \\
    Alice's Mother -- Alice's Sister 2 \\
    Alice's Mother -- Alice's Sister 3 \\
    Alice's Father -- Alice \\
    Alice's Father -- Alice's Sister 1 \\
    Alice's Father -- Alice's Sister 2 \\
    Alice's Father -- Alice's Sister 3 \\
    Alice's Mother -- Alice's Maternal Aunt \\
    Alice's Mother -- Alice's Maternal Uncle 1 \\
    Alice's Mother -- Alice's Maternal Uncle 2 \\
    Alice's Father -- Alice's Paternal Uncle \\
    Alice's Paternal Uncle -- Alice's Cousin \\
    \hline
    \textbf{First Round Verification and Generation} \\
    \hline
    Alice's Maternal Uncle 1 -- Maternal Cousin 1 \\
    Alice's Maternal Uncle 1 -- Maternal Cousin 2 \\
    Alice's Maternal Uncle 2 -- Maternal Cousin 3 \\
    \hline
    \textbf{Second Round Verification and Generation} \\
    \hline
    Alice's Paternal Uncle's Wife -- Paternal Nephew/Niece 1 \\
    \hline
    \textbf{Third Round Verification and Generation} \\
    \hline
    \st{Alice's Paternal Uncle's Wife -- Paternal Nephew/Niece 1}  \\
    Alice's Paternal Uncle -- Paternal Cousin 2 \\
    (inferring from the paternal side to meet the nephew/niece count) \\
    \textbf{The graph passes verification.} \\
    \hline
    \textbf{Question: How many cousins does Alice's sister have?} \\
    \hline
    Maternal Cousins: 3 (1 from Maternal Uncle 1 and \\
    2 from Maternal Uncle 2) \\
    Paternal Cousins: 2 (2 from Paternal Uncle) \\
    \textbf{Total Cousins:} \\
    3 (Maternal) + 2 (Paternal) = 5 Cousins \\
    \hline
\end{tabular}
}
\vspace{-0.2in}
\end{table}

\subsubsection{Analysis}
Based on the case studies, the effectiveness of \method may stem from its ability to infer missing entities and relationships. To validate this assumption, we designed an experiment where we manually added the missing roles and relationships in the AIW+ problem. Specifically, we completed the graph by incorporating the missing relationships. For example, we add \textit{One of Alice's mother's brother has 1 son while another has 2 sons. Alice's father also have another brother who has 1 son.} to the AIW+ example. We refer to this dataset as AIW+ Complete.


The results are shown in Table~\ref{table:complete}. From these results, we observe that all models, except for LLaMA 3.1-8B, perform well on this dataset. The performance difference between the AIW+ and AIW+ Complete datasets demonstrates that missing entities and relationships in the questions are a major barrier to LLM reasoning. The proposed \method addresses this issue by inferring the missing entities and relationships during the graph verification and generation processes, thereby improving performance. Additionally, \method continues to improve performance on the AIW+ Complete dataset, demonstrating that explicit graph structures can assist LLMs with this task.

\begin{table}[!htb]
\centering
\caption{The results of AIW+ Complete dataset.}
\vspace{-0.1in}
\label{table:complete}
\resizebox{\linewidth}{!}{
\begin{tabular}{c|cccc}
\hline
 & Claude & GPT-4o & Llama3.1-8B & Llama3.1-70B \\ \hline
Vanilla & 0.5882 & 0.8823 & 0 & 0.7058 \\
CoT & 0.5294 & 0.9411 & 0 & 0.8823 \\
\method & 0.7058 & 1 & 0.058 & 1 \\ \hline
\end{tabular}
}
\vspace{-0.2in}
\end{table}

\begin{figure}[!htb]
    \centering
    \vspace{-0.1in}
    \begin{subfigure}{0.49\linewidth}
        \centering
        \includegraphics[width=\linewidth]{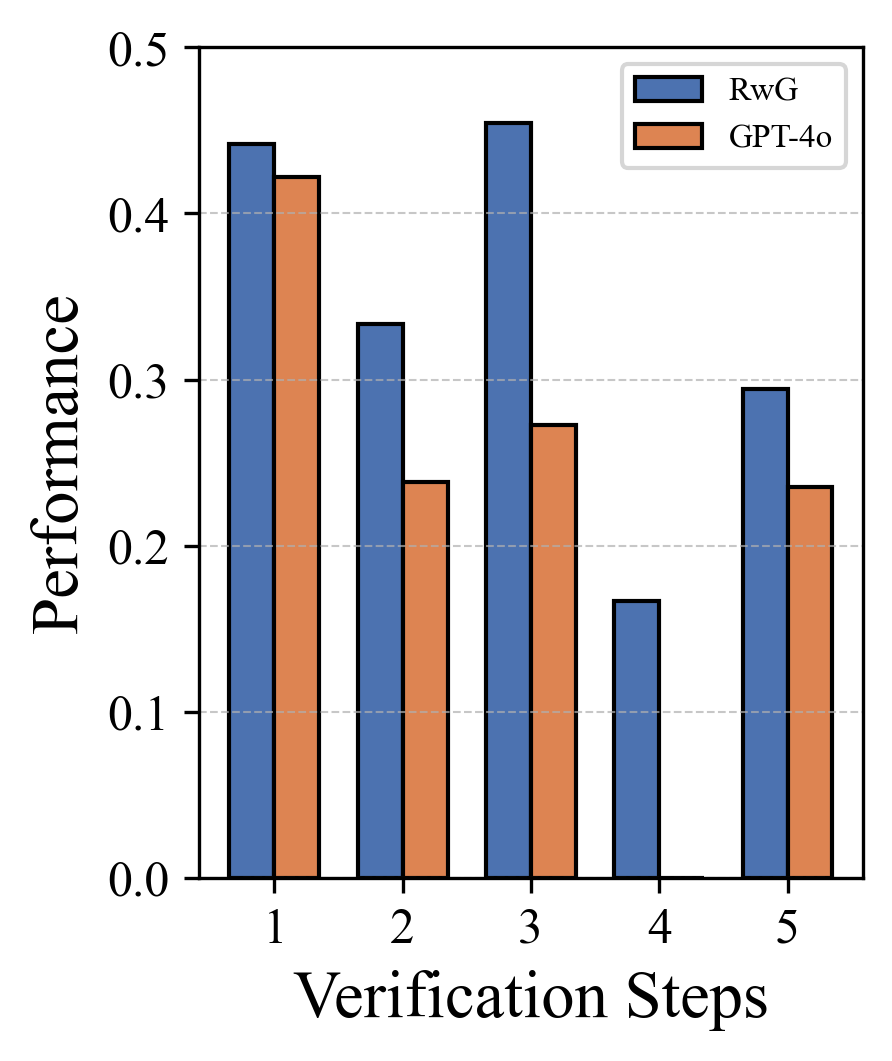}
        \caption{GPT-4o}
        \label{fig:subfig1}
    \end{subfigure}
    \begin{subfigure}{0.49\linewidth}
        \centering
        \includegraphics[width=\linewidth]{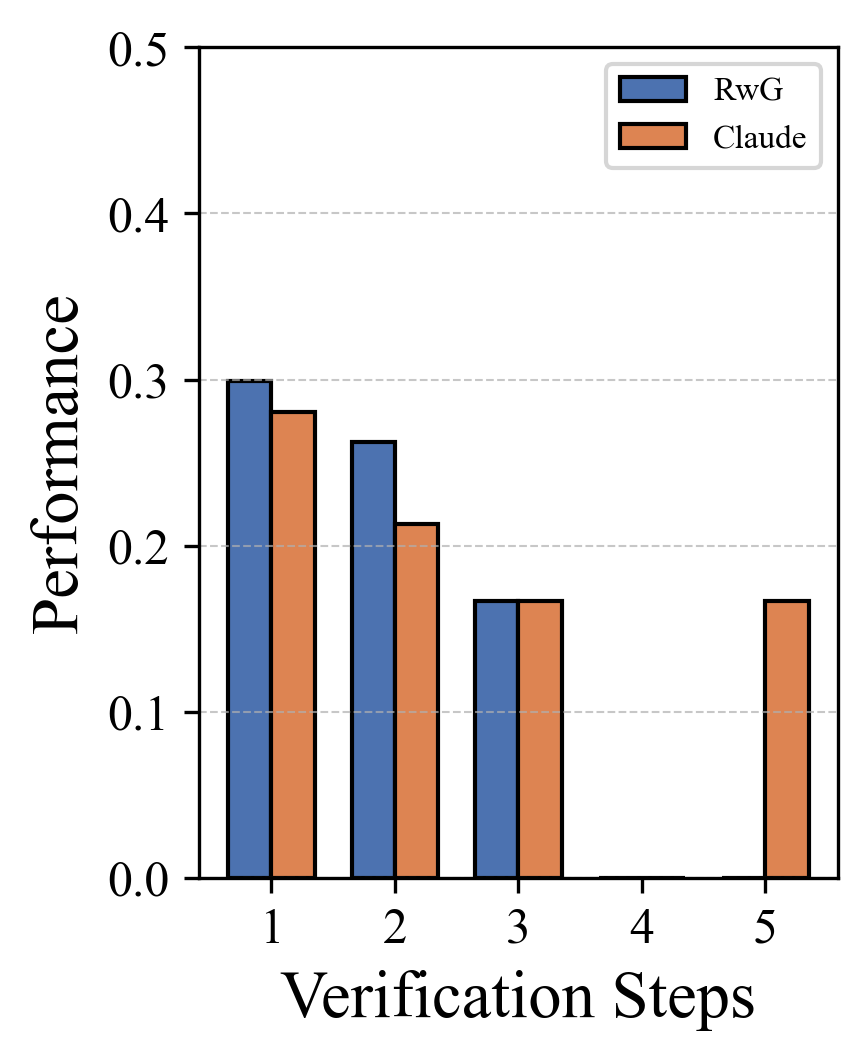}
        \caption{Cluade}
        \label{fig:subfig2}
    \end{subfigure}
    \vspace{-0.2in}
    \caption{Comparison of performances under different verification steps.}
    \label{fig:performance_gain}
    \vspace{-0.1in}
\end{figure}

We further analyze the performance gain of the proposed \method with respect to the number of verification steps. The number of verification and generation steps required to obtain the final graph varies depending on the question. If a question contains most of the entities and relationships, fewer verification steps are needed to construct the final graph. In contrast, if many verification and generation steps are required, the question is likely missing many entities and relationships, making it more difficult to solve.
We select the AR-LSAT dataset and compare the performance of the proposed \method with  vanilla models, as shown in Figure~\ref{fig:performance_gain}. 

We observe that when the verification step is 1, meaning the initial graph passes verification, the performance gap between \method and the vanilla models is small. However, as more generation steps are required to pass verification, the performance gap increases, which aligns with our assumption.

\begin{table*}[htb]
\centering
\caption{Comparison of different models on the Multi-hop Question Answering datasets}
\label{tab:multi-hop}
\resizebox{\linewidth}{!}{
\begin{tabular}{c|cc|cc|cc|cc}
\hline
Dataset & \multicolumn{2}{c|}{Hotpot} & \multicolumn{2}{c|}{MuSiQue} & \multicolumn{2}{c|}{2WikiMultihopQA} & \multicolumn{2}{c}{Clutrr} \\ \hline
Multi-hop QA & Claude & GPT-4o & Claude & GPT-4o & Claude & GPT-4o & Claude & GPT-4o \\
Vanilla & 0.700 & 0.7219 & 0.5008 & 0.6131 & 0.6608 & 0.8493 & 0.2488 & 0.5485 \\
CoT & 0.6941 & 0.7294 & 0.5492 & 0.6064 & 0.8160 & 0.8660 & 0.3721 & 0.6594 \\
Least-to-Most & 0.6943 & 0.7452 & 0.5799 & 0.6331 & 0.8160 & 0.8859 & 0.3385 & 0.6503 \\
Structure Prompting & 0.6547 & 0.7435 & 0.5594 & 0.6094 & 0.7594 & 0.8660 & 0.3834 & 0.6413 \\
\method & \textbf{0.7399} & \textbf{0.7742} & \textbf{0.6395} & \textbf{0.7187} & \textbf{0.8202} & \textbf{0.9040} & \textbf{0.4558} & \textbf{0.6911} \\ \hline
\end{tabular}
}
\vspace{-0.2in}
\end{table*}

\subsection{Task 2: Multi-hop Question Answering}

Multi-hop question answering typically provides several paragraphs of knowledge and requires answering a question that involves a sequence of interdependent reasoning steps leading to the final answer. These reasoning steps and their dependencies can often be represented as a directed acyclic graph (DAG). Therefore, the proposed \method aims to extract such reasoning graphs from given context to answer the multi-hop question. Since the given context can be lengthy and LLMs struggle to comprehend large graphs~\cite{dai2024revisiting}, we build only a subgraph related to the question rather than constructing the entire graph. The general generation and verification process for the multi-hop question answering task in \method is as follows: (1) \textbf{Generation:} Generate a graph related to the question by updating the previous graph, inferring missing relations, or adding more entities and relations from the context. (2) \textbf{Verification:} Verify whether the graph contains enough information to answer the multi-hop question.

\subsubsection{Datasets}

We selected four widely used multi-hop reasoning datasets: 2WikiMultihopQA~\citep{ho2020constructing}, MuSiQue~\citep{trivedi2022musique}, HotpotQA~\citep{yang2018hotpotqa} and Clutrr~\cite{sinha2018compositional}. More details are shown in Appendix~\ref{app:multi_dataset}

\subsubsection{Baselines}

We evaluate the proposed \method for multi-hop question answering using two LLMs: Claude 3-sonnet~\citep{anthropic2024claude} and GPT-4o~\citep{achiam2023gpt}. Additionally, we choose the following baselines: Chain-of-Thought (CoT)~\citep{wei2022chain}, Least-to-Most~\cite{zhou2022least}, Structure-Guided Prompting~\citep{cheng2024structure}. The detailed prompts can be found in Appendix~\ref{app:prompt}.

\subsubsection{Results}
\begin{figure}[!htb]
    \centering
   \includegraphics[width=0.9\linewidth]{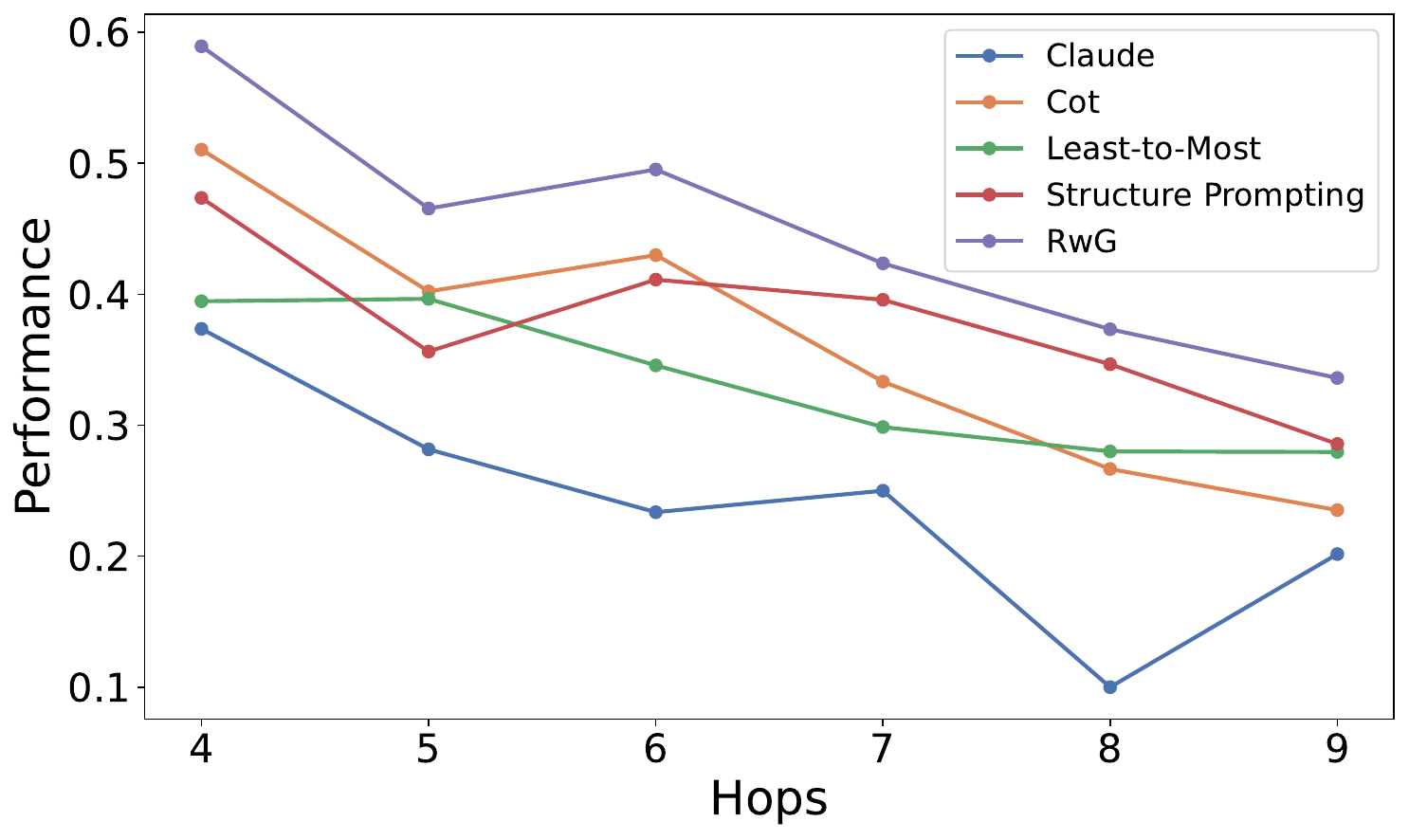}
     \vspace{-0.1in}
    \caption{Performance on different hop questions in the Clutrr dataset.}
    \label{fig:performance_clutrr}
\end{figure}

The overall performance on the selected datasets is shown in Table~\ref{tab:multi-hop}. Additionally, we evaluate the performance of different hop questions for the MuSiQue and Clutrr datasets, with results presented in Appendix~\ref{app:results}. Specifically, we illustrate the performance on different hop questions for the Clutrr dataset using Claude in Figure~\ref{fig:performance_clutrr}.
 \vspace{-0.1in}

From these results, we can make the following observations:
\begin{itemize}[leftmargin=0.2in]
\setlength\itemsep{-0.2em}
    \item The proposed \method outperforms all baselines on these representative multi-hop question answering datasets.
    \item The performance of all methods tends to decrease as the number of hops increases in both the MuSiQue and Clutrr datasets. However, the proposed \method performs well across both low and high hop questions.
    \item  Chain-of-Thought (CoT) tends to perform well when the number of hops is low, but its performance declines with higher hop questions, especially when using Claude. 
\end{itemize}

\subsubsection{Case studies}
In this section, we aim to understand why the proposed \method improves performance on multi-hop question answering tasks by analyzing several representative cases.

\textbf{Case 1:} We select an example from the MuSiQue dataset, where GPT-4o provided incorrect answers. In this example, 19 paragraphs are given, and the question is: \textit{When was the death penalty abolished in the country which, along with Eisenhower's VP's country, recognized Gaddafi's government early on?} The graph constructed by GPT-4o is as follows in Table~\ref{tab: case1}:
\begin{table}[ht]
    \centering
    \caption{The illustration of Case 1}
    \vspace{-0.1in}
    \label{tab: case1}
    \resizebox{\linewidth}{!}{
    \begin{tabular}{l}
    \hline
(Dwight D. Eisenhower) --[was President of]--> (United States) \\
(Richard Nixon) --[was Vice President of]--> (United States) \\
(United States) --[recognized government of]--> (Muammar Gaddafi) \\
(United Kingdom) --[recognized government of]--> (Muammar Gaddafi) \\
(Capital Punishment) --[abolished in]--> (Great Britain) [in 1965] \\
(Capital Punishment) --[abolished in]--> (Northern Ireland) [in 1973] \\
(Capital Punishment) --[abolished in]--> (United Kingdom) [in 1998] \\ \hline
    \end{tabular}
    }
    \vspace{-0.1in}
\end{table}

This graph provides a condensed representation of the entire context, containing essential reasoning structure to answer the question. By focusing on this explicit knowledge graph, which includes only the necessary information, LLMs can avoid irrelevant details that might otherwise interfere with their response generation~\cite{shi2023large}.

\begin{figure}[!htb]
    \centering
   \includegraphics[width=\linewidth]{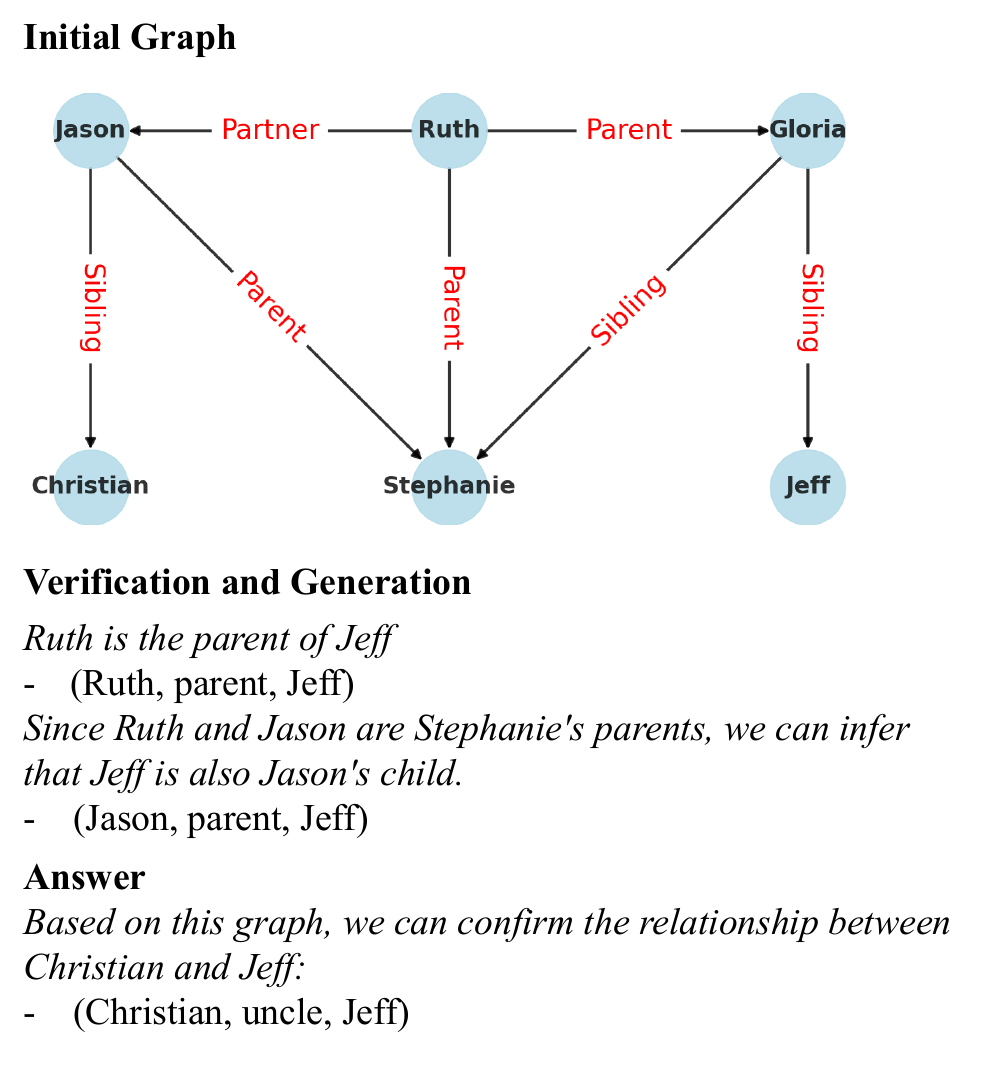}
   \vspace{-0.3in}
    \caption{The illustration of Case 2.}
    \label{fig:clutrr_example}
    \vspace{-0.1in}
\end{figure}

\textbf{Case 2:} We also select one example from the Clutrr dataset, which requires LLMs to infer multi-hop family relationships between \textit{Christian} and \textit{Jeff}. The procedure of \method with GPT-4o is shown in Figure~\ref{fig:clutrr_example}.  Although several methods leverage path information to infer relationships between entities, LLMs' ability to detect paths diminishes as the path length increases~\cite{dai2024revisiting}.  In contrast, the proposed \method would infer missing relationships during the graph construction.  For example, after the first round of verification and generation, the LLMs inferred an edge between \textit{Jason} and \textit{Jeff}, reducing the reasoning path length between \textit{Christian} and \textit{Jeff} from 4 to 2.

More examples can be found in Appendix~\ref{app:case}. Based on the case studies, the proposed \method aids LLMs in multi-hop question answering from two key perspectives: (1) The constructed graph reduces irrelevant information while maintaining an explicit reasoning structure; (2) The graph shortens the reasoning path length for the question.

\section{Conclusion}
\label{sec:conclustion}
\vspace{-0.1in}
In this paper, we propose a novel Reasoning with Graphs (\method) method to structure implicit knowledge to enhance the reasoning  capabilities of LLMs. Our method constructs graphs through multiple rounds of generation and verification, leveraging these graphs to answer complex questions. We evaluate our approach on both logical reasoning and multi-hop question-answering tasks using several widely recognized datasets. Experimental results demonstrate that \method significantly improves the performance of various LLMs across both tasks.

\section{Limitations}
In this paper, we aim to improve the reasoning ability of LLMs by modeling input as a graph structure, which mirrors the way humans often approach reasoning tasks. We conducted experiments on four popular LLMs: GPT-4o, Claude, Llama 3.1-8B, and Llama 3.1-70B. However, more LLMs can be tested with the proposed \method in future studies. Additionally, while we explored why explicit graph structures can aid LLM reasoning primarily through experimental results and case studies, a more rigorous theoretical analysis is an interesting direction for future work. Furthermore, our evaluation focused on logical reasoning and multi-hop question answering tasks, but other tasks can also be explored to assess the broader applicability of \method.

\bibliography{custom}

\clearpage
\newpage
\appendix
\section{Appendix}

\subsection{Datasets}
In this section, we introduce the used datasets in the logical reasoning task and multi-hop question answering task.
\subsubsection{Logical Reasoning Task}
\label{app:logical_dataset}
For the logical reasoning task, we select 4 datasets, i.e., AIW, AIW+~\citep{nezhurina2024alice}, LogiQA~\cite{liu2020logiqa} and AR-LSAT~\citep{wang2022lsat}. The details of each dataset are as follows:

\textbf{AIW:} AIW dataset contains a set of ``Alice in Wonderland Problems", which typically follow the format:  ``Alice has $N$ brothers and she also has $M$ sisters. How many sisters does Alice’s brother have?". This dataset is popular to evaluate the reasoning abilities of LLMs. 

\textbf{AIW+:} The AIW+ dataset is an extension of the AIW problem, describing a more complex family structure. It introduces additional hierarchy and distractors when depicting relational family structures, making the reasoning task more challenging. In the AIW+ problem, multiple solutions could arise if the model assumes that Alice's parents have additional children, which is also a feasible solution. To eliminate this ambiguity, we added a constraint to the problem: Alice's parents do not have any other children.

\textbf{LogiQA:} LogiQA is a widely used logical reasoning dataset that includes questions involving various types of reasoning, such as categorical reasoning, sufficient conditional reasoning, necessary conditional reasoning, disjunctive reasoning, and conjunctive reasoning. The dataset is divided into training, validation, and test sets. Since we do not train or fine-tune the LLMs, we selected 100 samples from the test set to evaluate different methods.

\textbf{AR-LSAT:} AR-LSAT is a dataset collected from the Law School Admission Test (LSAT). There are three dominant game types in LSAT: ordering games, grouping games, and assignment games. In ordering games, participants must be ordered based on given facts and rules. Grouping games involve separating participants into groups according to specific facts and rules. Assignment games require assigning characteristics to participants, such as scheduling tasks for individuals, while adhering to given rules. We use all the test data to evaluate the proposed method and baselines.

\subsubsection{Multi-hop Question Answering Task}
\label{app:multi_dataset}
For the multi-hop question answering task, we select 4 widely used datasets, i.e., 2WikiMultihopQA~\citep{ho2020constructing}, MuSiQue~\citep{trivedi2022musique}, HotpotQA~\citep{yang2018hotpotqa} and Clutrr~\cite{sinha2018compositional}. The details of each dataset are as follows:

\textbf{2WikiMultihopQA: } The 2WikiMultihopQA dataset is built from Wikipedia and Wikidata. It contains several related paragraphs and one question, with various types of multi-hop questions such as comparison, inference, compositional, and bridge-comparison questions. We randomly sampled 100 questions from the test set of 2WikiMultihopQA for the experiments.

\textbf{MuSiQue: } The MuSiQue dataset contains multi-hop questions via single-hop question composition. Like 2WikiMultihopQA, it includes several related paragraphs per question. The dataset features 2-hop, 3-hop, and 4-hop questions. For each hop type, we randomly sampled 100 questions. Detailed results for each hop can be found in Appendix~\ref{app:results}.

\textbf{HotpotQA:} HotpotQA is a widely used multi-hop question dataset. It provides 10 paragraphs to answer a single question. There are different difficulty levels, and the easier questions are typically solvable by LLMs. We randomly selected a subset of 100 hard bridging questions from the development set of HotpotQA.

\textbf{Clutrr:} The Clutrr (Compositional Language Understanding with Text-based Relational Reasoning) dataset differs from the other three multi-hop question datasets. It primarily contains a single paragraph that describes relationships between family members, and the task is to infer the relationship between two specified members. The dataset includes different path lengths between the predicted family members. For our experiments, we selected path lengths from 4 to 9, as shorter paths are generally easier for LLMs to solve. We report the overall performance in Table~\ref{tab:multi-hop} while the detailed results for each hop can be found in Appendix~\ref{app:results}.

\subsection{Prompts of \method}
\label{app:prompt}

In this section, we provide the prompt of the proposed \method for logical reasoning and multi-hop question answering tasks. The system prompt of the proposed \method is shown in Tabel~\ref{app:tab:system}.

\subsubsection{Logical Reasoning}

There are mainly three steps in the proposed \method, i.e., initial graph generation, graph verification and graph generation. During the experiments, we merge the graph verification and graph generation into one prompt for convenient. The initial graph generation prompt is shown in Table~\ref{app:tab:logical_inital}. The graph verification and generation prompt is shown in Table~\ref{app:tab:logical_generation}, and the question answering prompt is shown in Table~\ref{app:tab:logical_answer} .

\begin{table*}[!htb]
    \centering
    \caption{System prompt of \method.}
    \label{app:tab:system}
    \begin{tabular}{p{\textwidth}}
    \hline
You are an expert of knowledge graphs. Knowledge Graphs aim to represent the relationships between entities. You are good at reasoning based on the graph. When you are asked to output the graph, please write [latest graph] first, followed by all triples, such as (entity1, relation, entity2) in the graph. \\
    \hline
    \end{tabular}
       \vspace{0.2in}
\end{table*}

\begin{table*}[!htb]
    \centering
    \caption{Initial graph generation prompt for logical reasoning task.}
    \label{app:tab:logical_inital}
    \begin{tabular}{p{\textwidth}}
    \hline
    Please build a knowledge graph based on the given context: \{Context\} and question: \{Question\}. The graph aim to answer the question. The nodes represent entities while edges represent their relations. \\
    \hline
    \end{tabular}
\end{table*}

\begin{table*}[!htb]
    \centering
    \caption{Graph verification and generation prompt for logical reasoning task.}
    \label{app:tab:logical_generation}
    \begin{tabular}{p{\textwidth}}
    \hline
    Does the latest graph meet all the requirements? Please first define each relationships in the context. And then carefully verify all the requirements. If the old graph meets all the requirements, please write [YES] at the end. If the old graph is wrong, please update the graph by inferring missing relations and nodes -> Write the [latest graph] with new edge list first, followed by [No].
\\
    \hline
    \end{tabular}
    \label{tab:my_label}
\end{table*}

\begin{table*}[!htb]
    \centering
    \caption{Answer generation prompt for logical reasoning task.}
    \label{app:tab:logical_answer}
    \begin{tabular}{p{\textwidth}}
    \hline
 Please answer the following question based on the latest graph and context: \{Question\}.
\\
    \hline
    \end{tabular}
    \vspace{0.2in}
\end{table*}

\begin{table*}[!htb]
    \centering
    \caption{Initial graph generation prompt for multi-hop question answering task.}
    \label{app:tab:multi_inital}
    \begin{tabular}{p{0.2\textwidth}|p{0.8\textwidth}}
    \hline
    2WikiMultihopQA, MuSiQue, \qquad \qquad HotpotQA & There are multiple paragraphs in the given context: \{Context\}. Please first find all paragraphs that may related to the question: \{Question\}. Please extract all the entities and relations of these paragraphs. Then build a knowledge graph based on these entities and relations. \\ \hline
    Clutrr &   Please build a family relation knowledge graph based on the context sentence by sentence. Nodes represent roles, and edges represent relationships. The graph should be bidirectional, including ([entity1], relation, [entity2]) and ([entity2], reverse relation, [entity1]).  \\

    \hline
    \end{tabular}
\end{table*}

\begin{table*}[!htb]
    \centering
    \caption{Graph verification and generation prompt for multi-hop question answering task.}
    \label{app:tab:multi_generation}
    \begin{tabular}{p{0.2\textwidth}|p{0.8\textwidth}}
    \hline
    2WikiMultihopQA, MuSiQue, \qquad \qquad HotpotQA &  Does the graph include all the entities and relations related to the questions: \{Question\}? Please recursively add new entities and relations after you have new entities. If the old graph meets the requirement, please write [YES] at the end. If the old graph can not, please update the graph by retrieving more entities and relations from the given contexts. Add these information to form a new graph. -> Write the [latest graph] with new edge list first, followed by [No].  \\ \hline
     Clutrr & Can the [latest graph] contains enough information to answer the question: \{Question\}? Please confirm your conclusion. If yes, please write [YES] at the end. If not, update the graph by inferring missing relations between entities as many as possible based on the graph to form a new graph. Then, provide the [latest graph], followed by [No]. Please think step by step and explain your reasoning. \\
    \hline
    \end{tabular}
\end{table*}

\begin{table*}[!htb]
    \centering
    \caption{Answer generation  prompt for multi-hop question answering task.}
    \label{app:tab:multi_answer}
    \begin{tabular}{p{\textwidth}}
    \hline
 Please answer the following question: \{Question\} based on the latest graph and context: \{Context\}.\\ \hline
    \end{tabular}
\end{table*}

\subsubsection{Multi-hop Question Answering}
The prompts for Multi-hop Question Answering are similar to those used for logical reasoning. However, during graph generation, we only extract the entities and relationships relevant to the question to prevent the graph from becoming too large. During the verification stage, we check whether the current graph contains all the necessary information to answer the question. Specifically, the graph initialization prompt is shown in Table~\ref{app:tab:multi_inital}, the graph verification and generation prompt is shown in Table~\ref{app:tab:multi_generation}, and the question answering prompt is shown in Table~\ref{app:tab:multi_answer} .

\subsection{More results}
\label{app:results}
We provide detailed results for the different hop questions in the MuSiQue and Clutrr datasets. Specifically, the MuSiQue dataset contains 2, 3, and 4-hop questions, and the results are shown in Table~\ref{app:tab:musique}. Additionally, the results for 4 to 9-hop questions in the Clutrr dataset are provided in Table~\ref{app:tab:clutrr}. From the results, we observe that the proposed \method outperforms all baselines across all hop levels.
\begin{table*}[htb]
\centering
\caption{The results of different hop questions in MuSiQue dataset}
\label{app:tab:musique}
\begin{tabular}{c|cc|cc|cc}
\hline
Hops & \multicolumn{2}{c|}{MuSiQue 2} & \multicolumn{2}{c|}{MuSiQue 3} & \multicolumn{2}{c}{MuSiQue 4} \\ \hline
Methods & Claude & GPT-4o & Claude & GPT-4o & Claude & GPT-4o \\ \hline
Vanilla & 0.6011 & 0.7302 & 0.4937 & 0.5608 & 0.4076 & 0.5484 \\
Cot & 0.6472 & 0.7091 & 0.4967 & 0.5893 & 0.5039 & 0.5208 \\
Least-to-Most & 0.6900 & 0.7340 & 0.5222 & 0.6288 & 0.5275 & 0.5365 \\
Structure Prompting & 0.6325 & 0.7216 & 0.5572 & 0.5952 & 0.4886 & 0.5115 \\
\method & \textbf{0.7488} & \textbf{0.8126} & \textbf{0.6346} & \textbf{0.7032} & \textbf{0.5353} & \textbf{0.6403} \\ \hline
\end{tabular}
\end{table*}
\begin{table*}[htb]
\centering
\caption{The results of different hop questions in Clutrr dataset}
\label{app:tab:clutrr}
\resizebox{\linewidth}{!}{
\begin{tabular}{c|cc|cc|cc|cc|cc|cc}
\hline
Hops & \multicolumn{2}{c|}{Clutrr 4} & \multicolumn{2}{c|}{Clutrr 5} & \multicolumn{2}{c|}{Clutrr 6} & \multicolumn{2}{c|}{Clutrr 7} & \multicolumn{2}{c|}{Clutrr 8} & \multicolumn{2}{c}{Clutrr 9} \\ \hline
Methods & Claude & GPT-4o & Claude & GPT-4o & Claude & GPT-4o & Claude & GPT-4o & Claude & GPT-4o & Claude & GPT-4o \\ \hline
Vanilla & 0.3736 & 0.5263 & 0.2816 & 0.6091 & 0.2336 & 0.5981 & 0.2500 & 0.5902 & 0.1000 & 0.4800 & 0.2016 & 0.4873 \\
Cot & 0.5105 & 0.7894 & 0.4022 & 0.6839 & 0.4299 & 0.6915 & 0.3333 & 0.625 & 0.2666 & 0.5466 & 0.2352 & 0.5714 \\
Least-to-Most & 0.3947 & 0.7894 & 0.3965 & 0.7068 & 0.3457 & 0.7102 & 0.2986 & 0.5694 & 0.2800 & 0.5400 & 0.2796 & 0.5294 \\
Structure Prompting & 0.4736 & 0.7421 & 0.3563 & 0.7183 & 0.4112 & 0.6635 & 0.3958 & 0.6041 & 0.3466 & 0.5466 & 0.2857 & 0.5126 \\
\method & \textbf{0.5894} & \textbf{0.8315} & \textbf{0.4655} & \textbf{0.7471} & \textbf{0.4953} & \textbf{0.7102} & \textbf{0.4236} & \textbf{0.6597} & \textbf{0.3733} & \textbf{0.5533} & \textbf{0.3361} & \textbf{0.5798} \\ \hline
\end{tabular}
}
\end{table*}

\subsection{Case Studies}
\label{app:case}
In this section, we present additional cases generated by GPT-4o. Due to the lengthy context generated by GPT-4o, we have removed some non-essential content.

\subsubsection{Failure case of AIW+ problem with Chain-of-Thought prompting}
\label{app:case1}
In section~\ref{sec:case}, we analyzed how the proposed \method solves the AIW+ example. Here, we present a failure case using GPT-4o with Chain-of-Thought prompting.

\textbf{Question:} Alice has 3 sisters. Her mother has 1 sister who does not have children - she has 7 nephews and nieces and also 2 brothers. Alice's father has a brother who has 5 nephews and nieces in total, and who has also 1 son. Alice's parents have no other children.
How many cousins does Alice's sister have? Please think step by step and explain your reasoning.

\textbf{Response of GPT-4o with CoT:} \\
To solve this problem, we need to determine the number of cousins Alice and her sisters have. We'll approach this step by step:

1. Alice's Family Structure: \\
   - Alice has 3 sisters. \\
   - Alice's parents have no other children, so Alice and her 3 sisters are the only children in their immediate family. \\

2. Maternal Side: \\
   - Alice's mother has 1 sister and 2 brothers. \\
   - The mother's sister has no children. \\
   - The mother's 2 brothers collectively have 7 nephews and nieces. \\
   - Since Alice and her 3 sisters (4 children in total) are part of these 7 nephews and nieces, the remaining 3 must be Alice's cousins from her maternal uncles. \\

3. Paternal Side: \\
   - Alice's father has 1 brother. \\
   - This brother has 5 nephews and nieces in total. \\
   - Alice and her 3 sisters are 4 of these nephews and nieces. \\
   - Therefore, there must be 1 more nephew or niece, who is Alice's cousin from her paternal uncle. \\
   - This paternal uncle also has 1 son, who is Alice's cousin. \\

4. Total Cousins: \\
   - Alice has 3 cousins from her maternal side. \\
   - Alice has 1 cousin from her paternal side. \\
   - Therefore, Alice has a total of 3 + 1 = 4 cousins. \\

From the response, we can see that GPT-4o fails to infer one cousin from the paternal side, highlighting the effectiveness of the proposed \method.

\subsubsection{Case from LogiQA dataset with \method}

\textbf{Context}: There are nine cards, one to nine. A, B, C, D four people take cards, each take 2. It is known that the sum of the two cards taken by A is 10; the difference between the two cards taken by B is 1; the product of the two cards taken by C is 24; and the quotient of the two cards taken by D is 3.

\textbf{Question:} Which of the following is the left card? 0. Six.; 1. Three.; 2. Seven.; 3. Four.

\textbf{Initial Graph Generation}
The possible cards assignments:

A: (1, 9), (2, 8), (3, 7), (4, 6)

B: (2, 3), (3, 4), (4, 5), (5, 6), (6, 7), (7, 8), (8, 9)

C: (3, 8), (4, 6)

D: (3, 1), (6, 2), (9, 3)

Initial Graph: \\
- A - 1 ; A - 9 \\
- B - 8 ; B - 7 \\
- C - 4 ; C - 6 \\
- D - 3 ; D - 9 \\

\textbf{First Round Verification and generation}

There is overlap using card 9. New graph: \\
- A - 1 ; A - 9 \\
- B - 2 ; B - 3 \\
- C - 4 ; C - 6  \\
- D - 6 ; D - 2 \\

\textbf{Second Round Verification and generation}

Still found overlap. New graph: \\
- A - 1 ; A - 9 \\
- B - 4 ; B - 5 \\
- C - 3 ; C - 8 \\
- D - 6 ; D - 2 \\

\textbf{Third Round Verification and generation} 

Pass the verification.

\textbf{Answer: The left card is 7.}

\subsubsection{Case from AR-LSAT dataset with \method}
\textbf{Context:} A maintenance company that takes service requests from three clients—Image, Solide, and Truvest—plans to set targets for its average service response times. Service targets will be set at 3 days, 2 days, or 1 day. Two service targets are set for each client—one for requests received through the maintenance company's website and one for requests received by voicemail. The six targets are set according to the following conditions: None of the clients can have a website target that is longer than its voicemail target. Image's voicemail target must be shorter than the other clients' voicemail targets. Solide's website target must be shorter than Truvest's website target.

\textbf{Initial Graph Generation} \\
1. Identify the variables: \\
   - There are six targets:  \\
     - Image's website target (IW) and voicemail target (IV). \\
     - Solide's website target (SW) and voicemail target (SV). \\
     - Truvest's website target (TW) and voicemail target (TV). \\
The generated graph: \\
- IW $\leq$ IV \\
- IV < SV \\
- IV < TV \\
- SW < TW \\
- TW < TV \\
- SW < TV \\

\textbf{Question:} If Truvest's website target is shorter than its voicemail target, which one of the following must be true?
0. Image's voicemail target is 2 days.; 1. Image's website target is 2 days.; 2. Image's website target is 1 day.; 3. Solide's website target is 2 days.; 4. Solide's website target is 1 day.
\textbf{Answer:} \\
- Option 0: Image's voicemail target is 2 days. \\
      This option doesn't directly relate to the constraints provided. The constraints don't specify a numeric value for IV, just that IV < SV and IV < TV. Therefore, this option is not necessarily true based on the graph.

- Option 1: Image's website target is 2 days. \\
      Similar to Option 0, there is no specific constraint that pins IW to a numeric value. The constraint IW $\leq$ IV doesn't specify that IW must be 2 days. Hence, this option is not necessarily true.
- Option 2: Image's website target is 1 day. \\
      If IW is 1 day, it satisfies IW $\leq$ IV. However, there’s no direct constraint requiring IW to be exactly 1 day. Therefore, this option is not necessarily true.

- Option 3: Solide's website target is 2 days. \\
      For SW to be 2 days, while it satisfies SW < TW, the constraints don't dictate that SW must be exactly 2 days. No direct necessity is derived from the constraints for SW to be 2 days. Thus, this option is not necessarily true.

- Option 4: Solide's website target is 1 day. \\
      Given SW < TW and SW < TV, if SW is set to 1 day, it is possible to meet the condition of SW being the shortest among SW, TW, and TV, considering that TW > SW and TV > SW because of the constraints. Setting SW to 1 day ensures that it is less than TW and TV, which aligns with the necessary condition SW < TW and SW < TV. This option aligns well with the constraints and scenario.

\subsubsection{Case from 2WikiMultihopQA dataset with \method}
In the 2WikiMultihopQA dataset, most questions require only 2 hops, making it easier to extract the relevant entities in a single step.

\textbf{Question:} Which film has the director born earlier, Haiducii (Film) or My Wife'S Best Friend?

\textbf{Initial Graph Generation} \\
- (Haiducii, directed\_by, Dinu Cocea) \\ 
- (My Wife's Best Friend, directed\_by, Richard Sale) \\
- (Dinu Cocea, birthdate, 22 September 1929) \\
- (Richard Sale, birthdate, December 17, 1911) \\

\textbf{Answer:} My Wife's Best Friend

\subsubsection{Case from MuSiQue dataset with \method}
In the MuSiQue dataset, we selected a 4-hop question answering example for this case study.

\textbf{Question:} Based on population alone, what is the ranking of the city having the Courthouse of the birthplace of James Glisson in the country winning the rugby 7 in Las Vegas?

\textbf{Initial Graph Generation:} \\
- (James Glisson) --born in--> (Jackson County, Florida) \\
- (Jacksonville, Florida) --located in--> (Jackson County, Florida) \\
- (Bermuda) --participated in--> (Las Vegas 7's) \\
- (Las Vegas, Nevada) --hosted--> (Las Vegas 7's) \\
- (Jacksonville, Florida) --is ranked as most populous in--> (Florida) \\

\textbf{First Round Verification and Generation} \\
- (James Glisson) --born in--> (Jackson County, Florida) \\ 
- (Jacksonville, Florida) --located in--> (Jackson County, Florida)\\
- (Bermuda) --won--> (Las Vegas 7's) \\
- (Las Vegas, Nevada) --hosted--> (Las Vegas 7's) \\
- (Jacksonville, Florida) --is ranked 12th most populous city in--> (United States) \\

\textbf{Answer:} 12th

\end{document}